%% file: main.tex
\documentclass[10pt,twocolumn,letterpaper]{article}

\usepackage[pagenumbers]{cvpr} 

\usepackage{graphicx}
\usepackage{amsmath}
\usepackage[overload]{empheq}

\input{preamble}

\definecolor{cvprblue}{rgb}{0.21,0.49,0.74}
\usepackage[pagebackref,breaklinks,colorlinks,citecolor=cvprblue]{hyperref}

\def\paperID{16487} 
\def\confName{CVPR}
\def\confYear{2024}

\title{Single View Refractive Index Tomography with Neural Fields}

\author{Brandon Zhao$^{1*}$\quad Aviad Levis$^{1}$ \quad Liam Connor$^{2}$ \quad Pratul P. Srinivasan$^{3}$\quad  Katherine L. Bouman$^{1,2}$\\
\\
$^1$Department of Computing and Mathematical Sciences, California Institute of Technology\\  $^2$ Department of Astronomy, California Institute of Technology\\ $^3$Google Research\\
{\tt\small $^*$byzhao@caltech.edu}
}
\begin{document}
\maketitle

\begin{abstract}

   Refractive Index Tomography is the inverse problem of reconstructing the continuously-varying 3D refractive index in a scene using 2D projected image measurements. Although a purely refractive field is not directly visible, it bends light rays as they travel through space, thus providing a signal for reconstruction. The effects of such fields appear in many scientific computer vision settings, ranging from refraction due to transparent cells in microscopy to the lensing of distant galaxies caused by dark matter in astrophysics. Reconstructing these fields is particularly difficult due to the complex nonlinear effects of the refractive field on observed images. Furthermore, while standard 3D reconstruction and tomography settings typically have access to observations of the scene from many viewpoints, many refractive index tomography problem settings only have access to images observed from a single viewpoint. We introduce a method that leverages prior knowledge of light sources scattered throughout the refractive medium to help disambiguate the single-view refractive index tomography problem. We differentiably trace curved rays through a neural field representation of the refractive field, and optimize its parameters to best reproduce the observed image. We demonstrate the efficacy of our approach by reconstructing simulated refractive fields, analyze the effects of light source distribution on the recovered field, and test our method on a simulated dark matter mapping problem where we successfully recover the 3D refractive field caused by a realistic dark matter distribution.
\end{abstract}

\vspace{-.2in}
\section{Introduction}
\label{sec:intro}

\begin{figure*}
    \centering
    \includegraphics[width=\textwidth]{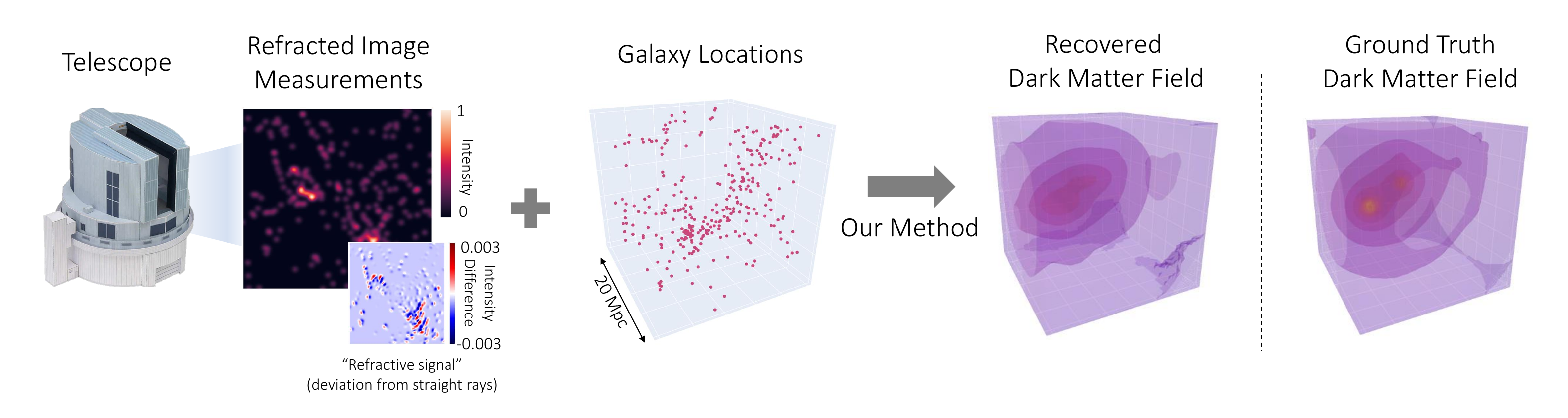}
    \caption{We formulate and present an approach for single-view refractive field reconstruction that explicitly models curved ray paths through a neural field. By leveraging spatial information and the emission profile of light sources located throughout the volume, we are able to localize refractive structures in 3D with just a single viewpoint. Here we demonstrate our approach on a simulation involving a realistic refractive field induced by simulated dark matter halos. Despite the large patches of space that are void of galaxy emitters, our method is still able to recover the rough structure of the 3D dark matter distribution from a single simulated telescope image of refracted galaxies. This reconstruction is also especially challenging because of the extremely weak degree of refraction, as seen in the relative scale of the intensity difference plot; the refractive effects of dark matter result in an intensity change of at most $\pm 0.3 \%$ of the maximum intensity of the original image. This experiment uses a realistic dark matter distribution derived from IllustrisTNG~\cite{ nelson2015illustris}, but we leave the inclusion of realistic measurement noise to future work. }
    \label{fig:teaser}
    \vspace{-.2in}
\end{figure*}

When light passes from one medium to another, such as from air to water, it encounters a change in \textit{refractive index} (RI). As a result, light rays are bent, or refracted, according to the magnitude of this change, causing effects like the shimmering (caustics) at the bottom of a pool of water. Media such as hot gas or translucent cell samples have RIs that continuously vary in space, causing light to take paths curved by the gradient of the underlying RI field. Although the RI field is not directly visible, its effect on the scene's illumination provides clues to its structure. 

Spatially varying RI fields are present in a diverse range of scientific applications. For example, certain fiber optic cables are designed to smoothly refract light away from the cable's outside padding \cite{reed2002gradient}. In seismology, pressure waves from earthquakes are refracted continuously as they travel through ground structures of varying density \cite{lin2009eikonal}. In astrophysics, gravitational effects from massive dark matter structures lens the light from faraway galaxies resulting in observable effects in telescope images. When the degree of these gravitational lensing effects is relatively weak, they can be modeled as refraction \cite{hoekstra2008weak}. 


In this work, we seek to tackle the problem of single-view RI tomography: 3D reconstruction of RI fields where image measurements are available only from a single viewpoint. This problem setting occurs when using telescope images or a single camera on scenes in the wild, and poses several challenges. First, image measurements are produced by rays traveling along continuously curving trajectories through the scene, where curvature is due to the structure of the underlying RI field. Unlike in linear tomography problems, where the projection geometry is independent of the medium, in RI tomography the ray paths are dictated by the unknown medium properties themselves. Thus the measurements are a non-linear function of the targeted refractive field. Second, limiting image measurements to a single viewing angle makes the problem severely ill-posed. That is, there are many possible refractive fields which could potentially produce the same image measurements. However, by leveraging the location of light sources distributed throughout the scene we can better constrain the recovery.

We propose a novel approach using neural fields with continuous ray tracing that leverages the physics of light propagation to recover an underlying refractive field. In particular, we represent the spatially varying RI field with a coordinate-based neural network, which we optimize through a differentiable rendering procedure that traces paths of light through the refractive field to best reproduce measurements. Coordinate-based neural networks are advantageous for this problem because they provide a continuous, easily-differentiable representation of the refractive field as a function of space. We also find empirically that coordinate-based neural networks lead to better optimization behavior out-of-the-box than other representations such as linearly interpolated grids.

Our approach expands upon recent works in neural scene representation \cite{mildenhall2021nerf, levis2022gravitationally}.  While in these works ray paths through the scene are assumed to be known, our approach explicitly models and solves for unknown refracted ray paths. We first study the conditions under which single-view reconstruction is possible, then test our approach by analyzing its ability to recover weak refractive fields. Next, we analyze the sensitivity of our approach to the density of light sources in the scene. Finally, as shown in Fig.~\ref{fig:teaser}, we include a case study motivated by the application of weak lensing where our method recovers the refractive field generated by physically realistic simulations of dark matter. 

\vspace{-.05in}

\section{Related Work}

\textbf{Image-Based Refractive Field Recovery} The reconstruction of refractive objects and phenomena is an ongoing field of study, both in applications that make use of standard consumer cameras as well as scientific imaging applications. 
 In Background-oriented Schlieren (BoS) imaging, images taken of a background with and without refractive effects are correlated to estimate an optical flow describing a projected measurement of the refractive field \cite{atcheson2008time}. Combining these projected measurements from multiple viewpoints through tomography gives an estimate of the 3D gradients, from which the 3D field can be estimated from a Poisson solver. A similar type of tomography can be done to passively model 3D turbulence statistics by combining time resolved deflection measurements from multiple viewpoints \cite{alterman2014passive}. Optical flow estimates such as those in BoS can also be used to detect refractive ``wiggles" in fluid flow that can then be used to track the projected motion of the refractive media; by observing these wiggles with stereo cameras the fluid's depth can also be recovered \cite{xue2014refraction}. Fluid refraction from water can also be estimated by submerged cameras viewing the sun as a guiding point \cite{alterman2014stella}. 
 
While the aforementioned methods utilize the physics behind ray optics, they don't explicitly model the continuously curved paths that rays take through these refractive media during 3D reconstruction. A method using the refractive curvature of traced rays to supervise an interpolated grid model has been shown to be effective for reconstructing and designing refractive media~\cite{teh2022adjoint}. However, as with the previous methods, this method still relies on access to multiple viewpoints to reconstruct the refractive volume. In this work, we instead seek to tackle the challenge of accurately reconstructing 3D refractive fields from a single viewpoint.

\vspace{-.15in}
\paragraph{Applications} Recent works in optics have used differentiable rendering techniques to design refractive surfaces for various applications. The materials making up the lens typically have a constant RI, meaning light is refracted a discrete number of times at the interface between the lens and the surrounding air. Thus, these works solve for the optimal shape of the material rather than an RI field varying within the lens. Fast differentiable ray tracing techniques have been used for the design of a single lens to produce a target image \cite{li2021end}, as well as compound lenses for camera design \cite{tseng2021differentiable}. These types of lens designs are aimed at reproducing a target image or favorable image effects. While we also aim to optimize a refractive field to match a set of observed image measurements, we are interested in recovering the properties of the refractive medium itself.

In biomedical imaging, Optical diffraction tomography (ODT) uses the intrinsic optical variation in a sample to reconstruct its RI field \cite{choi2007tomographic, sung2009optical}. Holographic measurements are taken from multiple illumination angles with coherent light modeled by wave optics. For example, in \cite{chowdhury2019high}, the authors present a technique to recover RI by using intensity-only measurements from multiple viewpoints. DeCAF \cite{liu2022recovery} uses a neural field representation to solve this inverse problem. Our problem setup differs from these biomedical applications in that we attempt to solve this problem without access to active illumination and coherent light sources.  Additionally, we use ray optics rather than wave optics in our characterization of light transport. 

We are motivated in particular by the application of gravitational lensing, where light coming from small elliptical sources (galaxies) curves on its path to Earth due to an invisible dark matter distribution. In the weak lensing regime, the small ray deflections caused by dark matter can be well-approximated as a refractive effect \cite{hoekstra2008weak}, allowing us to model the large-scale structure of the dark matter distribution as an RI field. Existing works use a linear tomography approach similar to that of BoS \cite{atcheson2008time} by first computing a 2D projected map of mass density, then backprojecting into 3D to obtain a reconstruction. Because image measurements are limited to a single viewpoint, filtered backprojection-style methods \cite{simon2009unfolding, vanderplas2011three} are susceptible to heavy smearing along the line of sight, even in the absence of noise. In \cite{leonard2012compressed, leonard2014glimpse} a compressed-sensing approach is used that assumes the density field can be sparsely represented by a dictionary of dark matter halos. However, this assumption is only approximately true in practice; our method can instead reconstruct 3D RI fields induced by a general matter distribution.


\vspace{-.1in}

\paragraph{Coordinate-based Neural Representations} Recently coordinate-based neural networks have gained traction for applications in vision and graphics. These models typically parameterize a three dimensional field, such as a scene's radiance, with the weights of a multi-layer perceptron (MLP). Coordinate-based neural representations have been used in graphics to model shapes with occupancy fields \cite{chen2019learning}, as well as three-dimensional realistic scenes with neural radiance fields \cite{mildenhall2021nerf}. Both of these works rely on a fixed straight ray assumption to render scenes for training and inference. Coordinate-based MLPs have also been used in black hole imaging \cite{levis2022gravitationally} to reconstruct emission fields with curved ray paths dictated by general relativity; however, the paths taken by rays of light are calculated then fixed beforehand according to a specific model of gravitational lens. In contrast, our work seeks to optimize ray paths that are unknown and directly estimated during model training. 

Eikonal ray-tracing with neural fields has been used for refractive tomography in the context of novel-view synthesis \cite{bemana2022eikonal}. However, this work uses images of the scene from multiple viewpoints for supervision. Additionally, it focuses on recovering scenes which have a piecewise constant RI (e.g. a glass block with constant RI surrounded by air), where no light emission  occurs within the refractive medium. While it achieves impressive results for rendering images from novel viewpoints, the reconstructed refractive indices of these volumes fail to match the ground truth. In contrast, our method leverages visible light sources from within the refractive medium to enable accurate reconstruction of a refractive volume from only a single viewpoint.



\section{Forward: Image Formation Model}
\label{sec:methods}
This section details the components of the image formation (forward) model. First, we introduce the equations governing the refraction and intensity measurements of individual rays. Next, we detail the numerical integration used to model image intensity measurements at a sensor. 

\subsection{Refractive Ray Tracing}
Purely refractive media are translucent and affect light rays by curving their paths rather than modifying their intensity. Light travels more slowly through more optically dense material, causing it to bend towards an increase in optical density. The RI $\eta$ is a scalar used to quantify the optical density of a material relative to a vacuum; in natural settings, $\eta \geq 1$ as light cannot exceed its speed in a vacuum. In this work, we are interested in volumes with a continuously varying refractive index. The effect of a continuously varying RI $\eta$ on a light ray's position $x$ and direction $v$ is described by Hamilton's equations for ray tracing \cite{ihrke2007eikonal}:

\begin{align}[left=\empheqlbrace]
    \frac{dx}{ds} &= \frac{v(s)}{\eta(s)} \label{eq:hamilton1}\\
    \frac{dv}{ds} &= \nabla\eta(s)
    \label{eq:hamilton2}
\end{align}

\begin{figure}
    \centering
    \includegraphics[width=0.4\textwidth]{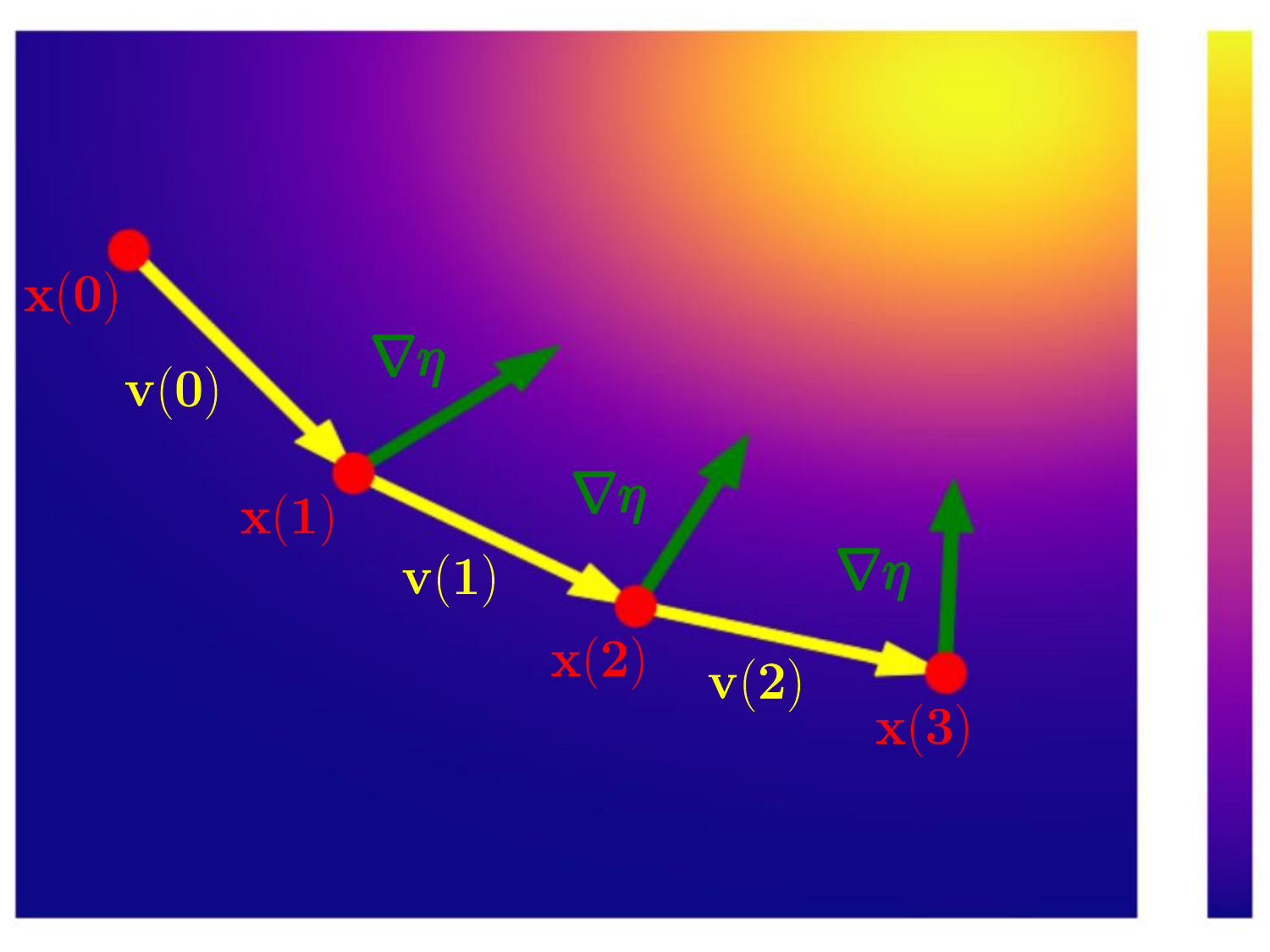}
    \caption{Illustration of ray tracing in a refractive field. Ray paths are calculated iteratively by solving the Hamiltonian equations \eqref{eq:hamilton1}, \eqref{eq:hamilton2} parameterized by the refractive field $\eta$. Rays are curved locally towards areas of high RI in the direction of $\nabla \eta$. }
    \label{fig:figure2}
\end{figure}

\begin{figure*}
    \includegraphics[width=\textwidth]{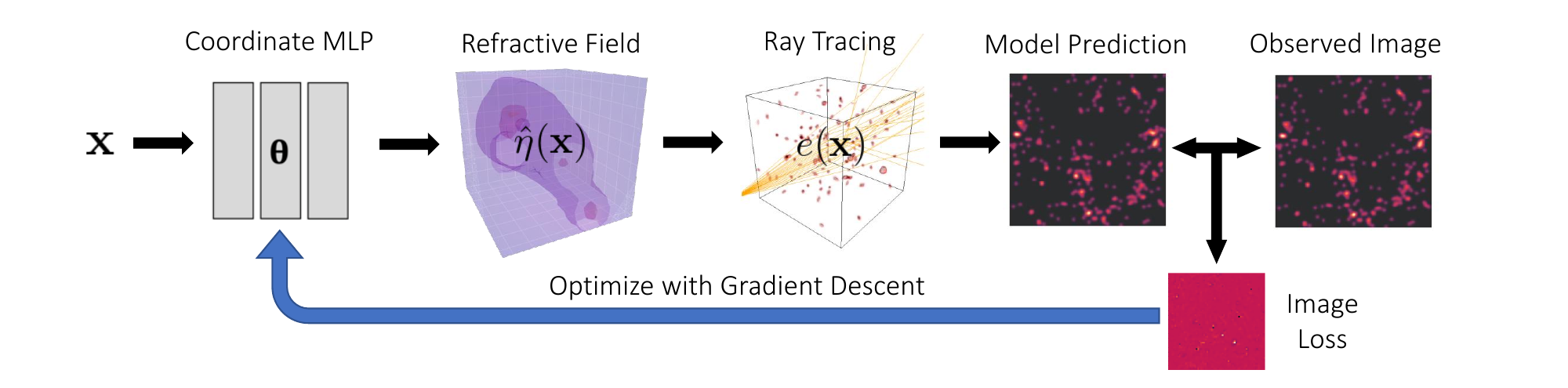}
    \caption{\textbf{Refractive Tomography Pipeline.} We model a 3D RI field $\hat{\eta}(\mathbf{x})$ as a continuous function using a neural network parameterized by $\mathbf{\theta}$. This refractive field, along with the known visible emission field $e(x)$ from point-like light sources, is fed through a ray tracing simulator to produce a predicted 2D image measurement. An image loss is then taken between the model and true image measurements to solve for $\mathbf{\theta}$. We differentiably trace rays through this refractive field and accumulate the light emitted along these curved ray paths to produce a 2D image. This allows us to optimize MLP parameters $\theta$ to minimize the loss between the rendered and true image measurements.}
    \label{fig: pipeline}
    \vspace{-.2in}
\end{figure*}
Here the derivatives of the 3D ray position $x$ and direction $v$ are taken with respect to the ray path length $s$, and are described in terms of the RI $\eta$ and its gradient $\nabla\eta$. The directional derivative of Eq. \eqref{eq:hamilton2} can be understood as an attractive force towards local increases in RI. 

Given an initial position and direction $x_0,v_0 \in \mathbb{R}^3$, equations \eqref{eq:hamilton1} and \eqref{eq:hamilton2} describe a first-order initial value problem whose solution  $ \Gamma(s) = (x, v)$ is a curve describing the trajectory of a ray as a function of its path length $s$: 

\begin{equation} \label{eq:raytrace}
    \Gamma(s) = (x_0, v_0) +  \int_{0}^{s} (x'(\tilde{s}), v'(\tilde{s})) \,d\tilde{s} 
\end{equation}
An illustration of refractive ray tracing is given in Figure \ref{fig:figure2}. 

\subsection{Sensor and Light Emission Integration}

The sensor is modeled as a single-wavelength pinhole camera positioned outside of the scene. The intensity image $I$ of a given scene is represented as a discrete function indexed by pixel coordinates $i,j$. The specific intensity of a pixel $I(i,j)$ is given by integration of light emission over a ray path solution $\Gamma_{i,j}$, to Eq. \eqref{eq:raytrace}. The initial conditions $\Gamma_{i,j}(0)$ are defined by the position and direction of a ray traveling backwards from the sensor through the pixel center.

Within the scene, light emission $e(\mathbf{x})$ is represented as a continuous scalar function of a 3D coordinate $\mathbf{x}=(x,y,z)$, and is assumed to be known. This assumption is commonly used for refractive modeling problems in the forms of calibration images in BoS imaging \cite{atcheson2008time} or galaxy catalogs in dark matter mapping \cite{hoekstra2008weak}. 
Given the ray path $\Gamma_{i,j}$ starting at the pixel in location $(i,j)$, we model the specific intensity $I(i,j)$  by taking the line integral of emission coefficients $e$ along the path $\Gamma_{i,j}$: 

\begin{equation} \label{eq:emission}
    I(i,j) = \int_{\Gamma_{i,j}} e(\mathbf{x}(s)) \,ds 
\end{equation}
Eq.~\eqref{eq:emission} corresponds to a simplified model of light transport in which the specific intensity is conserved along a ray's path. In gravitational lens theory \cite{schneidergravitational}, the emission term in the integrand is decreased for sources in areas of high refractive index due to gravitational redshift. However, for gravitational RI fields with $\eta$ close to 1, which we focus on in this work, this effect is negligible and we thus exclude it from our forward model~\cite{levis2022gravitationally}. In other applications of refractive tomography \cite{bemana2022eikonal, pediredla2020path}, the refractive radiative transfer equation (RRTE) \cite{ament2014refractive} may be more appropriate for the rendering forward model. Further details, including experiments with the RRTE, can be found in the supplement.

\subsection{Numerical Integration}

A ray path solution $\Gamma = (x,v)$ to the Hamiltonian initial value problem given by equations \eqref{eq:hamilton1} and \eqref{eq:hamilton2} is fully determined by the initial position and direction $\Gamma_{i,j}(0)$ of the ray traced through a pixel $(i,j)$, the refractive field $\eta$, and its gradient $\nabla \eta$. Thus, we rewrite the integral in Eq. \eqref{eq:raytrace} as: 
\begin{equation} \label{eq:solveham}
    \Gamma_{i,j} = \texttt{solveHam}(\Gamma_{i,j}(0), \eta, \nabla \eta ).
\end{equation}
Integrating the scene's emission function $e$ along the ray path $\Gamma$ as in Eq. \eqref{eq:emission} allows us to calculate image intensities: 
\begin{equation} \label{eq:render}
    I(i,j) = \texttt{renderIm}(\Gamma_{i,j}, e ).
\end{equation}
In general, closed-form solutions for integrals \eqref{eq:solveham} and \eqref{eq:render} do not exist, so they are solved numerically. We use an adaptive step-size Dormand-Prince integrator \cite{dormand1980family} to solve Eqs. \eqref{eq:solveham} and \eqref{eq:render} in our experiments. Backpropagating through the integrator using autograd requires memory in the order of the number of solver steps. Instead, we follow previous works \cite{bemana2022eikonal, teh2022adjoint, chen2018neural} in using the adjoint formulation \cite{pontryagin1987mathematical} which has constant memory requirements for backpropagation.

\section{Inverse: Estimating a Refractive Field}

In this work, we focus on weakly refractive fields where the deflections of a single source are on the order of fractions of a pixel. Although this may seem small, we rely on the combined signal across an ensemble of lensed sources. 

To solve the inverse problem of refractive tomography in this setting, we use an analysis-by-synthesis approach. We model the continuous refractive field with a coordinate-based MLP, which is optimized to fit a true observed image. In the following section we describe the neural representation and optimization scheme.  

\subsection{Continuous Refractive Field Representation}
To form a candidate continuously-varying 3D refractive field, we use a coordinate-based MLP inspired by the NeRF \cite{mildenhall2021nerf} model. The MLP, parameterized by weights $\mathbf{\theta}$, takes as input spatial coordinates $\mathbf{x}$ and outputs a scalar index of refraction. To ensure that our reconstruction is physically plausible, we apply the Softplus function $\sigma$ to restrict the output of our network to be greater than 1:
\begin{equation}
    \hat{\eta}_\theta(\mathbf{x}) = \sigma(\text{MLP}(\gamma(\mathbf{x})); \mathbf{\theta}).
\end{equation}
Here, $\gamma(\mathbf{x})$ is a positional encoding layer mapping an input coordinate to a set of Fourier basis functions \cite{tancik2020fourier} with exponentially-increasing frequencies: 
\begin{equation}
    \gamma(\mathbf{x}) = [\sin(\mathbf{x}),\cos(\mathbf{x}), \dots, \sin(2^{L-1}\mathbf{x}), \cos(2^{L-1}\mathbf{x})]^T.
\end{equation}
Adding this positional encoding allows the MLP to fit continuous fields with higher spatial frequencies as the degree $L$ is increased \cite{tancik2020fourier}. We find empirically that a degree of $L=4$ is suitable for naturally smooth refractive fields. 


Ray tracing through our model field requires integrating equation \eqref{eq:solveham} during each training iteration. This requires evaluating the model $\hat{\eta}_\theta$ and its spatial gradient $\nabla \hat{\eta}_\theta$ at continuous spatial locations over the course of optimization, and for both of these functions to be differentiable with respect to the model parameters $\theta$ for gradient descent. In other words, we need to be able to evaluate $\nabla_\theta \nabla_x \hat{\eta}_\theta$ at arbitrary points throughout the modeled refractive field. We use the eLU activation function rather than a ReLU so that this gradient is smooth during optimization, and can be easily evaluated with existing autograd frameworks. 


\subsection{Optimization}
Our neural field model defines a continuous refractive field $\hat{\eta}_\theta$ and its gradient $\nabla_x \hat{\eta}_\theta$ parameterized by its weights $\theta$. Thus, a rendered image $\hat{I}$ obtained by ray-tracing through a neural field $\hat{\eta}_\theta$ with Eqs. \eqref{eq:solveham} and \eqref{eq:render} is a function of $\theta$.

An overview of our pipeline is shown in Fig.~\ref{fig: pipeline}. To train the neural network weights $\theta$, we solve a minimization problem whose objective is a mean-squared error between the observed image $I(i,j)$, and the rendered image $\hat{I}(i,j; \theta)$, plus a regularization term $\mathcal{R}(\theta)$ to enforce a boundary condition of $\eta = 1$ on the cube's faces: 

\begin{equation}
    \min_\theta \mathcal{L}(\mathbf{\theta}) = \sum_{i,j}||I(i,j) - \hat{I}(i,j, \theta)||_2^2 + \lambda \mathcal{R}(\theta).
\end{equation}

This loss is fully differentiable with respect to model parameters $\theta$, allowing us to use gradient descent to solve the optimization. The term $\mathcal{R}(\theta)$ acts only as a constraint on the boundary conditions of the volume; unlike explicit regularizers, such as total variation, we rely on the implicit regularization from the neural field for our reconstruction. More details on the $\mathcal{R}(\theta)$ term can be found in the supplement.

\subsection{Implementation Details}

In our experiments, we use an MLP with 4 layers, where each layer is 256 units wide. The neural network is implemented in JAX \cite{jax2018github} and trained with the Adam optimizer \cite{kingma2014adam} with exponential learning rate decay from $10^{-4}$ to $5\times 10^{-6}$ over 10K iterations. The weights of the neural network were randomly initialized with the He uniform variance scaling initializer \cite{he2015delving}. Code implementation will be publicly available on the project page. 
\vspace{-0.2in}
\paragraph{Unmodeled Sources of Noise} Because our method leverages information about the location and shape of light sources to enable reconstruction from a single viewpoint, it is sensitive to noise in both our knowledge of light source locations and in image measurements. An analysis of our method's sensitivity to these noise sources can be found in the supplement. We plan to investigate ways to overcome this sensitivity in future work.

\begin{figure*}
    \centering
    \includegraphics[width=\textwidth]{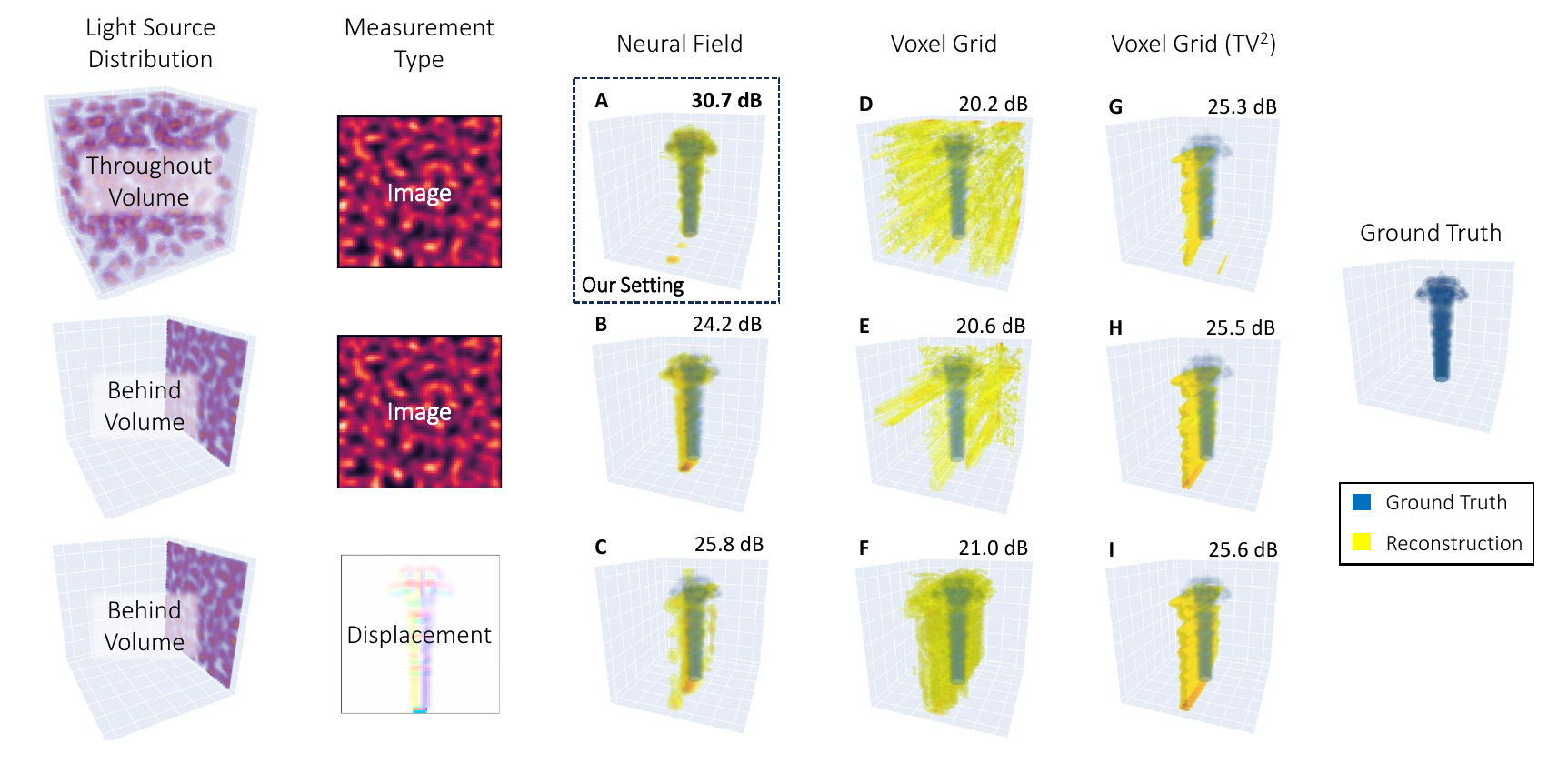}
    \caption{\textbf{Measurement and Reconstruction Conditions.} We compare our neural field reconstruction model to a linear voxel grid with and without TV$^2$ regularization for gas flow reconstruction under three different measurement conditions. While single-view recovery is difficult with measurements of ray displacement, we find that image measurements of light sources spread throughout the refractive volume makes recovery possible with our method. In addition, a neural field's implicit regularization results in better-constrained reconstructions than the linear voxel grid, even when an explicit regularizer is used. In each recovery we super-impose the ground-truth gas plume in blue over the reconstruction in yellow for comparison. We note that all reconstructions shown fit the measurements almost perfectly; differences in their structures highlight the ill-posedness of the problem.}
    \label{fig:gasflow}
    \vspace{-.2in}
\end{figure*}
\vspace{-.1in}
\section{Results}
\label{sec:experiments}

In this section we present results of our method on general refractive tomography problems. We generate ground truth refractive fields inside of a three-dimensional volume, where the RI takes a constant value of 1 outside the region of interest. Each ground truth refractive field is defined by evaluating a continuous function at regularly spaced grid points. The continuous field is then evaluated by linearly interpolating between these grid points. Image measurements are rendered using a pinhole camera with a $64^2$ resolution. Throughout this section, we quantify the accuracy of our reconstructions by sampling the ground truth and reconstructed fields on the same uniformly spaced grid of sample points as the ground truth and report the PSNR. When calculating the PSNR, we first subtract 1 from the RI. 


\subsection{Measurement and Reconstruction Conditions}

In this section we compare refractive reconstructions under nine conditions: three sets of measurement conditions paired with three different reconstruction strategies. We show that single-view RI tomography is made possible by a combination of the measurement conditions of our problem setting and the implicit regularization given by our neural field model. Results are shown in Fig.~\ref{fig:gasflow}.

\vspace{-.0in}

{\it Measurement Conditions:} The first measurement condition mirrors the astronomical domain that motivates our work, in which image measurements are taken of light sources distributed throughout the refractive volume (Fig.~\ref{fig:gasflow} top row). The second conditions we compare to are image measurements taken of light sources behind the volume (middle row). Third, we investigate using ray displacement measurements rather than image measurements; in this case, the displacements correspond to a plane of light sources behind the volume (bottom row). To simulate the displacement measurements, we ray trace through the ground truth refractive field to recover a perfect optical flow map at the same resolution as our sensor. Although this would be infeasible in practice, we use these measurements to represent the ideal limit of measurements from a single-view BoS setup \cite{atcheson2008time}. 

{\it Reconstruction Conditions:} We also compare reconstructions from three models: our neural field representation, a linear voxel grid, and a linear voxel grid with TV$^2$ regularization, with regularizer weight chosen to maximize PSNR with the ground truth volume. These correspond to the last three columns of Fig.~\ref{fig:gasflow}, respectively. The goal of the regularized voxel grid is to show that our method can outperform even the highest-performing TV$^2$ model; in practice, it would be infeasible to choose the regularizer weight in this way as we don't have access to the ground truth.
\vspace{-0.2in}
\paragraph{Data \& Results} In Fig.~\ref{fig:gasflow} we demonstrate the effect of each pair of conditions in reconstructing the fuel injection dataset (a $64^3$ RI grid) from SFB 382 of the German Research Council (DFG). Both a method using an unregularized voxel grid as well as BoS have been shown to reconstruct this volume given access to multiple viewpoints \cite{teh2022adjoint, atcheson2008time}.
We report the PSNR for each reconstruction, but note that it isn't fully representative of reconstruction quality, which should also be evaluated visually. 

{\it Effect of Measurement Conditions:} When optimizing our neural field model on image measurements of light sources located throughout the refractive volume, accurate 3D recovery of the gas flow is possible with minimal artifacts (A). However, in (B), where the light sources are instead located behind the volume, the neural field is able to recover the rough structure of the gas flow but not its absolute location.  When using displacement measurements, none of the reconstructions (C, F, I) are able to correctly recover the shape of the gas flow from a single viewpoint, even with a perfect optical flow map. While it has been shown that voxel grid optimization with displacement measurements can accurately reconstruct this volume \cite{teh2022adjoint}, having access to only a single viewpoint causes the reconstruction to be smeared along the optical axis. An analysis of the voxel grid recovery with access to additional viewpoints can be found in the supplement. We conclude that using image measurements of light sources spread throughout the volume is a key ingredient enabling 3D reconstruction from a single viewpoint.

\begin{figure}
    \centering
    \includegraphics[width=0.45\textwidth]{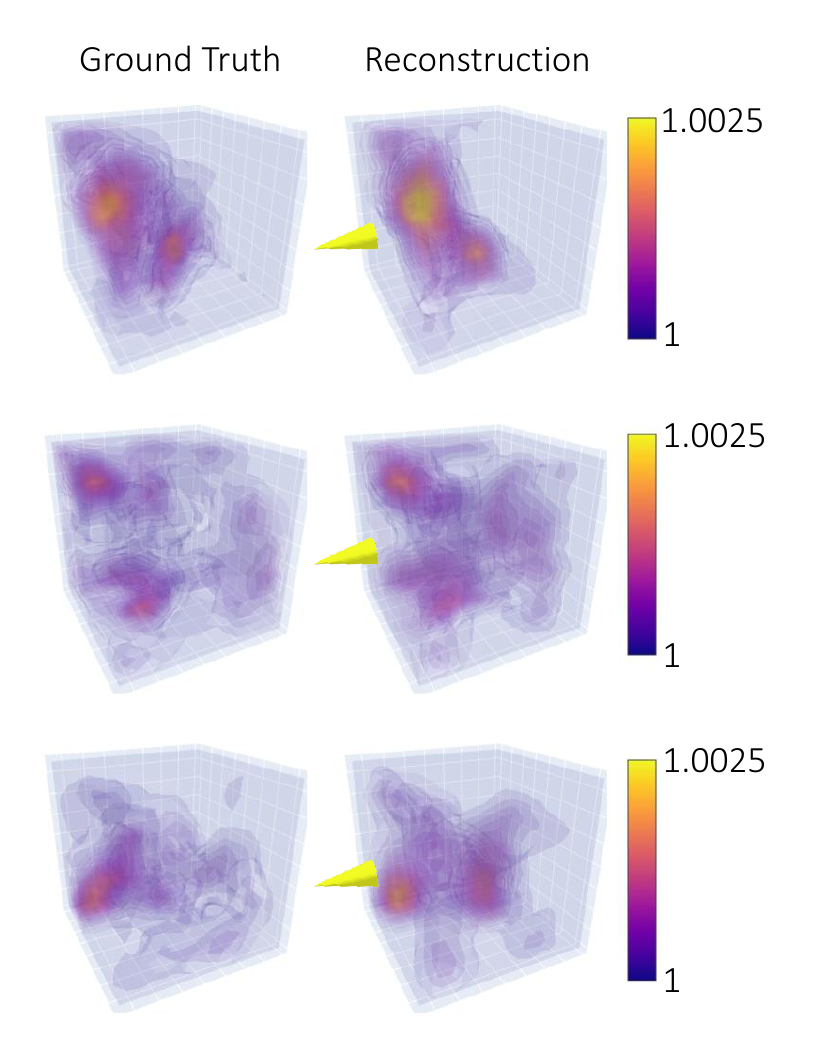}
    \caption{\textbf{Recovery of Smooth Refractive Fields.} Each row corresponds to a refractive tomography problem with a different random refractive field against a random distribution of 290 light sources. Although refractive recovery from a single viewpoint is highly ill-posed, our method can leverage source location information to detect and localize 3D refractive structure. However, as can be seen in the third example, due to the ill-posed and highly non-convex nature of this tomography problem, our model is still be susceptible to error. The yellow cones indicate sensor direction.}
    \label{fig:matern}
    \vspace{-.3in}
\end{figure}

{\it Effect of Reconstruction Conditions:} An image-based loss on point-like light sources is highly nonconvex, and thus models such as an unregularized voxel grid can be prone to converging to bad minima. Although all reconstructions in Fig.~\ref{fig:gasflow} fit the measurements almost perfectly, each converges to a different refractive field. The unregularized grid reconstructions (D, E) optimize locally along ray paths to fit the image measurements, resulting in streak-like artifacts and a highly non-smooth recovery. Even when an explicit TV$^2$ regularizer is used on the voxel grid, as in recoveries (G) and (H), it still fails to recover the structure of the gas flow along the optical axis. The implicit regularization of our neural field model appears to favor compact reconstructions, which matches the gas flow volume well. In the following section, we show that this implicit regularization also allows us to recover more general smooth fields.

\subsection{Recovery of Smooth Refractive Fields}
\label{sec:smooth}
We demonstrate our approach by recovering randomly generated smooth refractive fields in a dense field of light sources from a single viewpoint. We use Poisson Disk sampling \cite{bridson2007fast}
to generate an emission field of 290 light source locations.  
Light sources are amplitude scaled 3D Gaussians with independent random elliptical orientations.
Ground truth refractive fields are generated from a random Gaussian Process with a smooth covariance kernel described in the supplement. These true fields were then scaled to an RI range of 1 to 1.003.  To put this RI magnitude into context, the refractive effect of these fields causes a median deviation of 0.03 to 0.07 pixels on a $64^2$ pixel image for rays shot parallel through the volume from the front to the back. 

Fig.~\ref{fig:matern} shows that our method has the potential to accurately reconstruct complex, smooth fields. However, due to the ill-posedness and highly non-convex nature of this problem, our model is still susceptible to reconstruction artifacts, as seen in the last row. Note that no explicit regularization besides a boundary condition term was used in solving for these refractive fields. Thus, our results suggest that our neural network architecture combined with a low degree of positional encoding act as an implicit prior on the smoothness of our reconstruction, allowing it to effectively recover smooth refractive fields.

\begin{figure}
    \centering
    \includegraphics[width=0.45\textwidth]{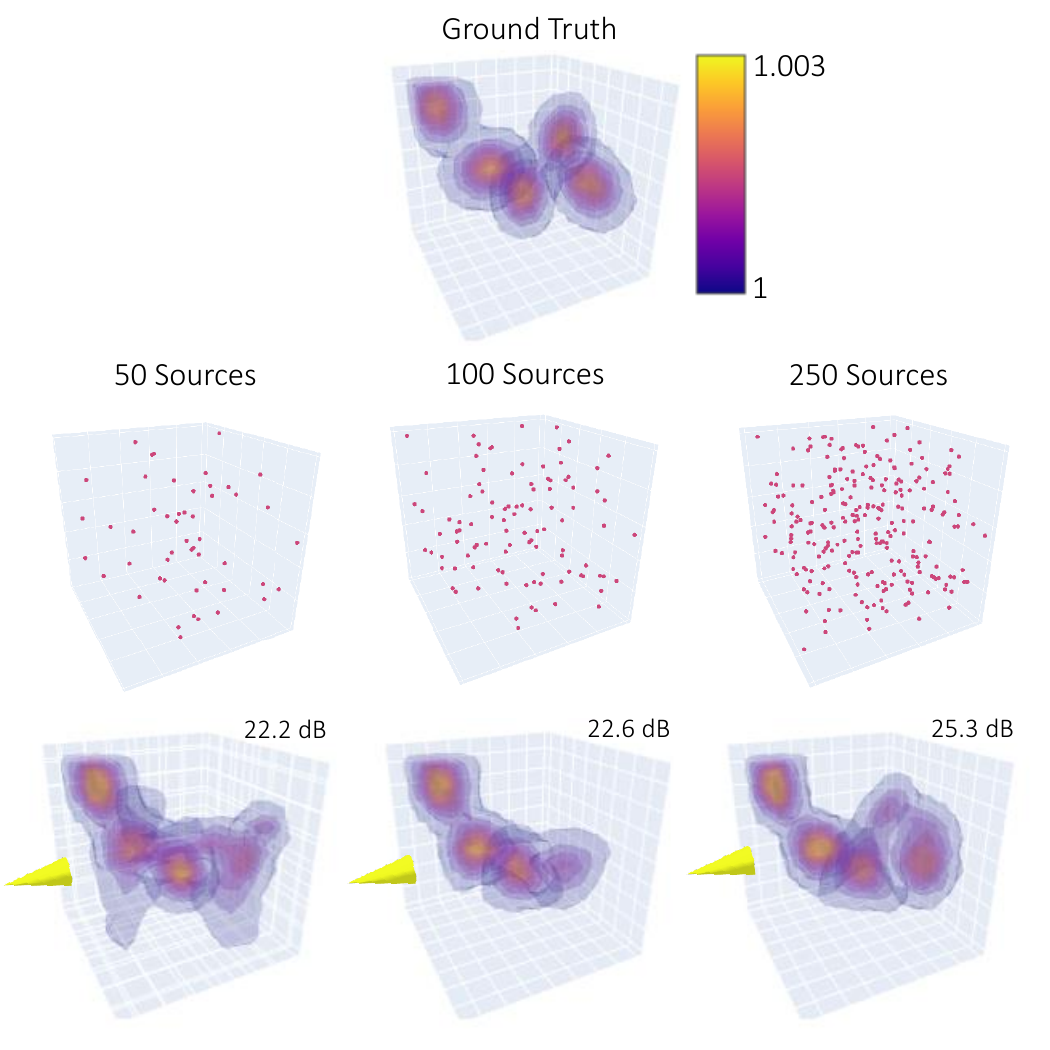}
    \caption{\textbf{Sensitivity to Light Source Density.} We analyze our approach's reconstruction of a refractive field given light source fields with a range of densities. As the number of light sources decreases, it becomes harder to detect and localize refractive objects, especially towards the back of the volume. However, even with sparse light sources, our method is able to accurately resolve refractive objects in the foreground. The yellow cones indicate sensor direction.}
    \label{fig:ls}
    \vspace{-.2in}
\end{figure}

\subsection{Sensitivity to Light Source Density} 
\label{sec:sparse}

From a single viewpoint, refractive fields are detected when they lie between light sources and the sensor. Background refractive structures become difficult to detect with a sparse array of light sources since fewer light sources lie behind them. Because we supervise our model with curved ray paths throughout the scene, our model uses the deflections between adjacent light sources to localize the refractive field. Therefore, we expect that as available light sources in the scene become sparser, our model will struggle to detect and localize the structure of a refractive field.

We perform a refractive field reconstruction on a scene containing five elliptical refractive objects in light source emission fields of varying densities. We sample 250 light source locations uniformly, and create emission fields of decreasing density by taking random subsets of this initial sample. We find that reconstruction quality of our approach degrades with sparser light source fields, especially in the background portion of the field. However, even with only 50 light sources in the volume, our method is able to accurately resolve the structure of refractive objects in the foreground. Results are shown in Fig.~\ref{fig:ls}.

\section{Case Study: Dark Matter Mass Mapping}
\label{sec:darkmatter}

In this section we show an application to astronomical imaging of dark matter from simulations. In outer space, clumps of invisible mass called dark matter permeate the cosmic web. These clumps, called halos, are so massive that their gravitational pull warps the surrounding space, which lenses nearby passing rays of light. In the weak lensing regime, the effects of gravitational lensing are well-described by refraction from an ``effective'' RI field \cite{narayan1996lectures}. In fact, the degree of curvature in weak lensing is so small that its effect on galaxy images is often modeled by a rotation (rather than a translation) of the galaxy's appearance on the sky \cite{bartelmann2001weak}. By combining telescope images with photometric redshift measurements of the distances of galaxies from Earth, scientists are able to determine their approximate spatial locations. The inverse problem of dark matter mass mapping is to reconstruct an invisible distribution of dark matter given refracted images from telescope surveys and photometric measurements of galaxy distances \cite{hoekstra2008weak}. 


\subsection{Simulated Dark Matter Halo Recovery}

We now use our method to recover a realistic dark matter refractive field. To generate this refractive field, we use the cosmological hydrodynamical simulation IllustrisTNG, which includes both dark and baryonic matter \cite{nelson2015illustris, pillepich2018simulating}. We use the catalog of dark matter halos produced by the TNG300-1 simulation, focusing on a $20^3$ Mpc$^3$ region containing 1747 massive halos. As galaxy placement is known to trace the dark matter distribution, we placed Gaussian emission functions emulating galaxies at the center of the 300 most massive halos. We aim to reconstruct a realistic dark matter distribution, but leave the inclusion of realistic measurement noise to future work. 

RI tomography in the weak lensing regime requires sensors with sufficient resolution to detect the small shearing effects from gravitational lensing. In our experiments, we use an increased sensor resolution of $512 \times 512$; this resolution in comparison with the scale of our scene is consistent with that of current and upcoming weak lensing surveys \cite{laureijs2011euclid, takada2010subaru}. We train our model using SGD, sampling batches of 4096 rays per iteration due to memory constraints. 

Our reconstruction of the invisible dark matter field is shown in Fig.~\ref{fig:teaser}. Our method is able to accurately recover the structure of the dark matter field with slight damping on the largest peaks of halo centers.
This recovery is especially challenging because of the distribution of galaxies throughout the scene; large patches of the image are dark as galaxies are not uniformly distributed throughout the volume. In addition, the amplitude of the refractive field generated by these halos is three orders of magnitude lower than in our previous results, resulting in a weak refractive signal; the median deflection caused by the refractive field is 6$\mathrm{e}{-3}$ pixels. We find that despite these challenges, our method is able to recover a refractive field that correlates strongly with true simulated dark matter halos, and shows promise for application to 3D dark matter recovery in the future.


\section{Conclusion}
\label{sec:disc}

In this paper we present a novel method for the tomography of continuous refractive fields from single-view image measurements with neural fields. Our method succeeds at reconstructing simulated refractive fields with varying light source density. By using the known location of light emitters, our method reconstructs refractive volumes from a single viewpoint that rival those of multiview methods. Finally, our results on physically realistic simulations of dark matter show promise for the application of our method to real weak lensing survey data for dark matter mapping. Beyond dark matter fields, our method opens doors to reconstructing other refractive phenomena, underscoring the computer vision community's potential to drive significant advances in scientific discovery with computational imaging.





{
    \small
    \bibliographystyle{ieeenat_fullname}
    \bibliography{main}
}

\end{document}



\maketitle

\hypersetup{linkcolor=black}
\tableofcontents
\pagebreak


\begin{figure*}
    \includegraphics[width=\textwidth]{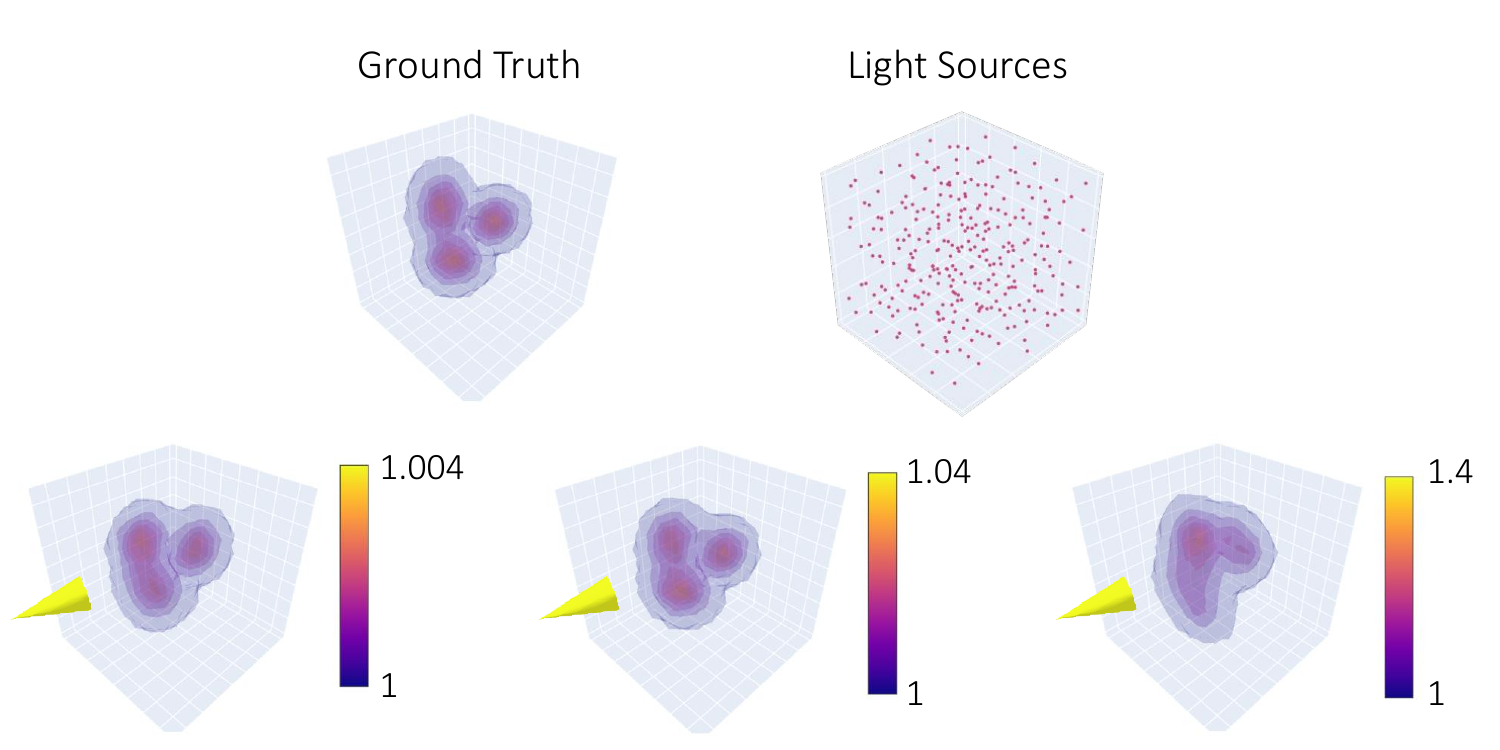}
    \caption{\textbf{Robustness to Strong Refraction}: Our approach explicitly models the continuous nature of refraction, and therefore curving ray paths, within an optimization framework. Using a continuous model that allows for curving ray paths enables robust reconstructions of refractive fields with a wide range of magnitude. We highlight how our approach is able to reconstruct the underlying refractive field across three orders of magnitude. As the magnitude of refraction increases the problem becomes more ill-posed due to the potential for large deviations in ray paths, leading to the observed reconstruction degradation in the high magnitude case. The yellow cones indicate sensor direction.}
    \label{fig:recon_mag}
\end{figure*}

\section{Robustness to Strong Refraction}

Typically, refraction tomography (e.g. Schlieren tomography~\cite{atcheson2008time}) recoveries rely on estimation of the total deflection of light rays entering and exiting a refractive volume. The underlying assumptions are that light is refracted once and that ray paths are approximately linear. Although these assumptions are valid for refractive fields with a refractive index (RI) concentrated at a small volume (e.g. transparent gas flows) they become inaccurate in large volumes with a high RI. In contrast, our method explicitly models the continuous curvature of ray paths throughout the volume. By doing so our approach is able to reconstruct strongly refractive fields where ray paths are continuously curving and do not adhere to the approximately linear model or single refraction. We illustrate the strength of our approach in its ability to recover a range of refractive fields with increasing magnitudes.  Figure \ref{fig:recon_mag} shows that our method is able to accurately recover the volume in all cases with minimal artifacts. However, it is important to note that as the magnitude of refraction increases the problem becomes more ill-posed due to the increased potential for crossing ray paths; this leads to some degradation of results when the magnitude of refraction is large.

\section{Sensitivity to Noise}

\begin{figure*}
    \includegraphics[width=\textwidth]{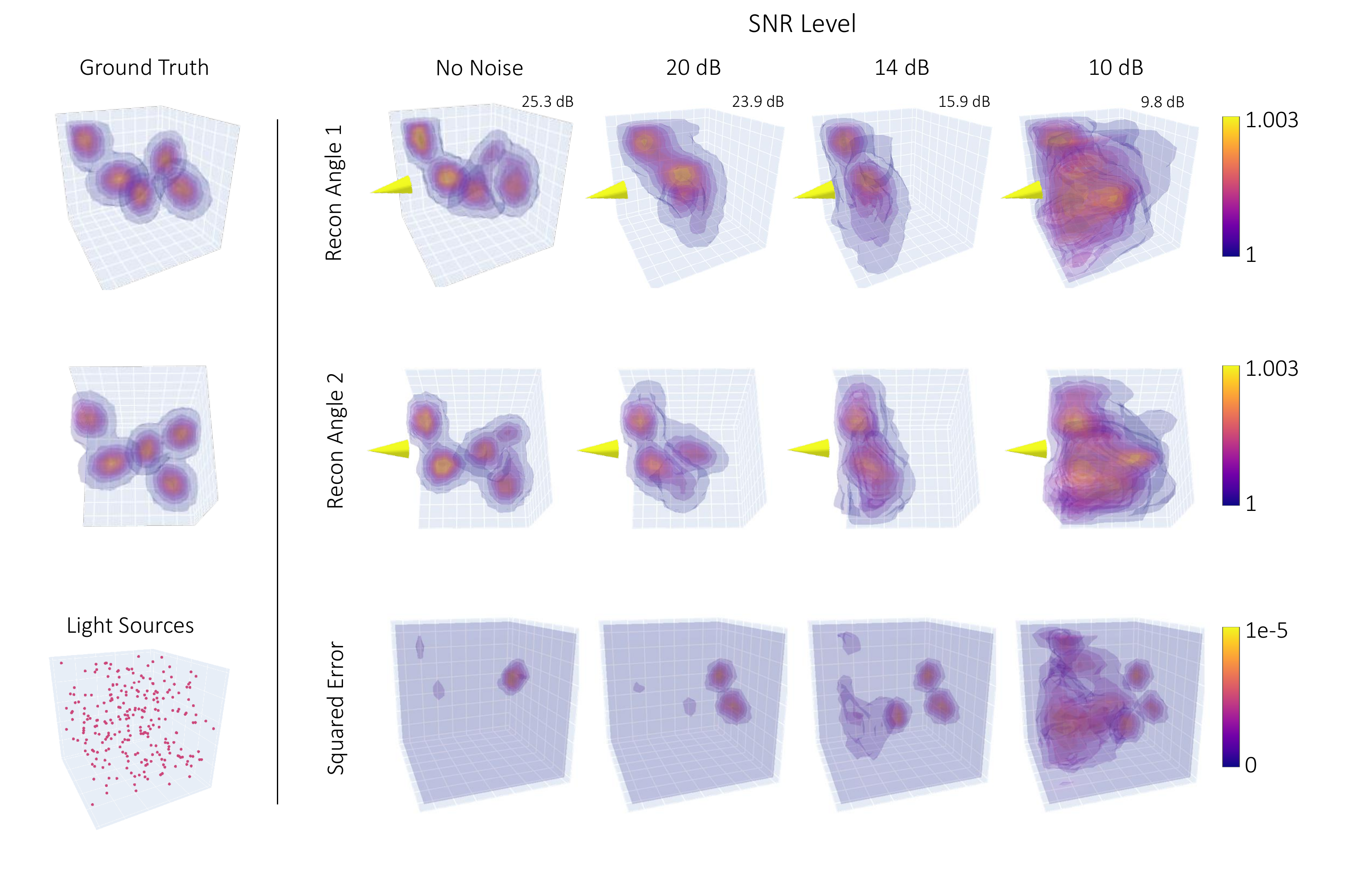}
    \caption{\textbf{Image-Level Noise}: 
    In these simulations, we highlight the effect of image measurement noise on the reconstruction of a weakly refractive field. The left column shows the ground-truth refractive field together with the distribution of light sources. Each reconstruction (each column) highlights the effects of a different measurement SNR level on the reconstructed volume. The volumetric error plot highlights the non-uniform effects of image noise resulting in recovery error increasing with distance from the sensor. Note that no explicit spatial regularization is being used in reconstruction. The yellow cones indicate sensor direction.}
    \label{fig:im_noise}
\end{figure*}

Our reconstruction method uses image measurements and knowledge of the locations and shapes of light sources to reconstruct the 3D refractive medium. In this section, we examine the effects of different noise sources. The first is additive pixel noise that affects image intensity measurements. The second is noise in our estimate of light source properties (position/orientation). We analyze the effects these noise sources would have on the reconstruction with a range of noise levels. 

\subsection{Image-Level Noise}
\label{subsec:image_noise}
To simulate sensor noise in our image measurements, we add white Gaussian noise to each pixel measurement. Reconstruction is performed on a RI field consisting of five elliptical Gaussian ``blobs''. Light sources are modeled as randomly oriented elliptical Gaussians uniformly distributed throughout the volume. Each light source emits a maximum of $\sim 1$ unit of radiance resulting in an image signal-to-noise ratio (SNR) of approximately: ${\rm SNR} = 10 \log(\sigma_{\rm noise})$. To prevent over-fitting to the measurement noise, we rely on early stopping. We stop optimization when the data fit term in the optimization loss reaches the expected noise mean square error (MSE). The underlying assumption of using this criterion is that we have a reasonable model for the additive sensor noise. 

In Figure \ref{fig:im_noise} we show results for three SNR levels: 20, 14, 10 [dB] which are typical for current observational instruments \cite{oguri2018two, mandelbaum2018first}. We highlight the impact of image noise on the reconstructions from two viewing angles (top rows), as well as the error volume comparing the recovery to the ground truth (bottom row). At higher SNR our approach is able to reconstruct several refractive Gaussian blobs with minor blurring artifacts. Lower SNR levels impact the recovered structures resulting in reconstruction artifacts. In this work we show a general reconstruction method as a first step that does not rely on explicit regularization. We expect that future applications incorporating specific informative priors could help to mitigate these noise artifacts. Furthermore, although many galaxies (light sources) lie within the SNR$\approx 10$ regime, there are many brighter or closer sources in a higher SNR regime that could be used for a reliable albeit partial reconstruction of the universe.  

\subsection{Uncertainty in Light Source Orientation}

\begin{figure*}
    \includegraphics[width=\textwidth]{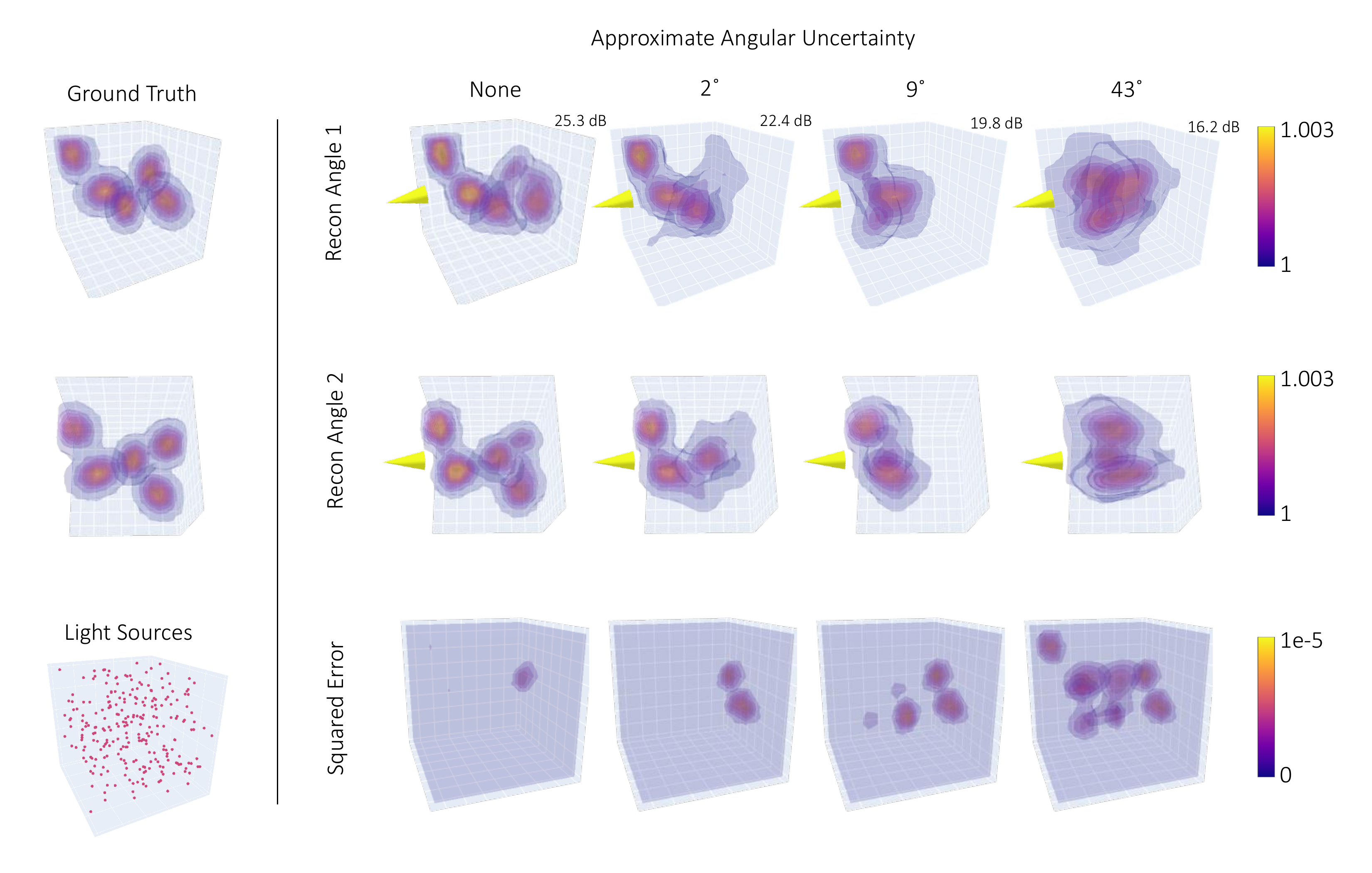}
    \caption{\textbf{Sensitivity to Noise in Light Source Orientation}: In these simulations, we highlight the effect of uncertainty in the orientation of each light source on the reconstruction. The left column shows the ground-truth refractive field together with the distribution of light sources. Each reconstruction (each column) corresponds to a different level of uncertainty in the light source orientation. Due to symmetry, a $90^\circ$ rotation is the maximum orientation difference. The volumetric error plot highlights the non-uniform effects of source orientation noise resulting in increasing recovery error with distance from the sensor. Note that no explicit spatial regularization is being used in reconstruction.The yellow cones indicate sensor direction.}
    \label{fig:ls_noise}
\end{figure*}

Our approach relies on knowing the shapes and positions of the light sources. In the astronomical setting, we have access to an estimated distance through spectral observations (further sources appear redder due to an effect called cosmological red-shift). Furthermore, weak refraction causes small distortions in the image plane that are best approximated as rotations of an elliptical source (rather than translations). Therefore, in this section, we analyze the 
 dominant noise source which comes from the uncertainty in the source orientation. 

To analyze this effect we reconstruct a refractive field (see Sec.~\ref{subsec:image_noise}) with randomly perturbed light source orientations. Each light source is randomly rotated in the image plane (about the depth axis) at an angle sampled from a mean-zero Gaussian distribution. The noisy orientations are subsequently used to generate observations for reconstruction. Analogously to the early stopping criteria described in Sec.~\ref{subsec:image_noise} we prevent over-fitting to noise by estimating the image noise induced by the noisy orientation. We estimate the expected rotation MSE by comparing 10 random sampled orientations to an unrotated image. This empirical MSE estimate is used as a stopping criterion for the optimization. 

In figure \ref{fig:ls_noise} we show results for three levels of uncertainty in the source orientation corresponding to an uncertainty of ${\sim} 2^\circ$, ${\sim} 9^\circ$, and ${\sim} 43^\circ$, where the uncertainty is given by a range of $\pm 3$ standard deviations of the orientation noise distribution. Fig.~\ref{fig:ls_noise} highlights that our approach is able to reconstruct the underlying structure well given good estimates of light source orientations. As expected, the reconstruction quality degrades with a degree of uncertainty. Note that again no explicit regularization is being used in these reconstructions. 


\section{Multi-View Recovery with Voxel Grids}
\begin{figure*}
    \includegraphics[width=\textwidth]{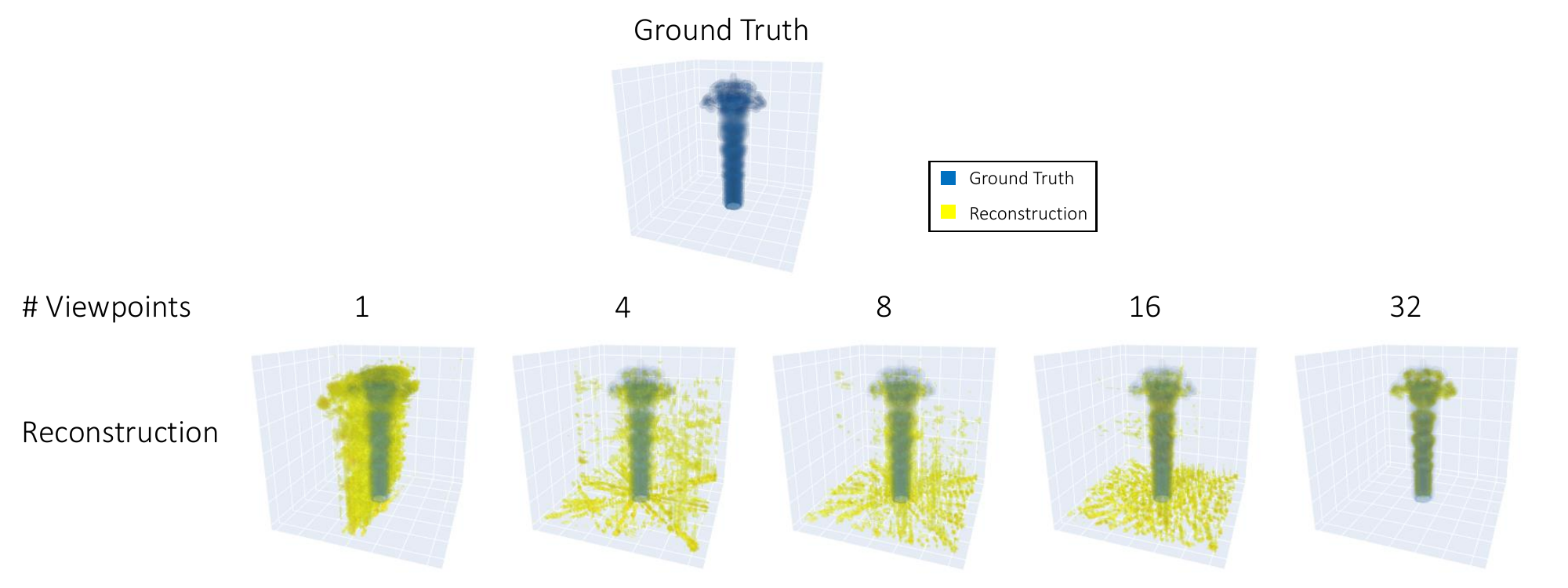}
    \caption{\textbf{Multiview Recovery with Voxel Grids}: In these simulations, we analyze the reconstruction performance of an unregularized voxel grid on a fuel injection RI volume with a varying number of viewpoints. With fewer than 32 viewpoints, reconstructions suffer from projection artifacts due to the underconstrained nature of the recovery. In particular, for a single viewpoint, the reconstruction is smeared along the optical axis. In each recovery we super-impose the ground-truth gas plume in blue over the reconstruction in yellow for comparison. All reconstructions shown fit the measurements almost perfectly; differences in the structures are due to the ill-posedness of the problem.}
    \label{fig:multiview}
\end{figure*}

In Sec. 5.1 of the main paper we highlight that single-view RI tomography is made possible by a combination of specific measurement conditions and our neural field reconstruction strategy. We show single-view reconstruction results on a synthetic fuel injection volume; both a method using an unregularized voxel grid as well as BoS techniques have been shown to reconstruct this volume given access to 32 viewpoints \cite{teh2022adjoint} and displacement measurements. In this section we reproduce  multi-view reconstruction results for voxel grid optimization on this volume, varying the number of viewpoints from single to multi-view. In particular, as the number of viewpoints is decreased, the voxel grid reconstruction suffers from projection artifacts as the problem is underconstrained. Only with a large number of viewpoints can the unregularized voxel grid reconstruct the volume accurately with few artifacts, which motivates the use of an alternate strategy for single-view recovery. Results are shown in Fig.~\ref{fig:multiview}.

\section{Boundary Regularization}
In our simulations, we reconstruct refractive volumes assuming a constant value of 1 outside of their boundary. In this way, we avoid modeling refraction outside of the volume of interest.  In our reconstructions, we regularize the predicted field boundary to have a value of 1 by minimizing the following loss function with respect to network parameters $\theta$: 
\begin{equation}
    \min_\theta \mathcal{L}(\mathbf{\theta}) = \sum_{i,j}||I(i,j) - \hat{I}(i,j, \theta)||_2^2 + \sum_{k=1}^n (\hat{\eta}_\theta (\mathbf{x}_k) - 1)^2.
    \label{eq:bc_reg}
\end{equation}
The second term in Eq.~\eqref{eq:bc_reg} enforces the model to output $1$ at points ${\bf x}_k$ chosen from a uniform grid on the volume boundary.

\section{Generating Random Refractive Fields}
In Section 4.1, we show reconstruction results on smooth, randomly generated refractive fields. We use a Gaussian random process to generate fields as follows: Let G be a mean-zero Gaussian process with the Matern kernel, with length scale 0.25 and smoothness parameter $\nu = 1.5$. We sample G at an evenly spaced $16 \times 16 \times 16$ grid of points within the unit cube, then pass these outputs through a soft-plus function to give non-negative values. Finally, we scale the soft-plus outputs, add 1, then pad these points at the cube boundaries with a constant value of 1. We then use trilinear interpolation to evaluate the refractive field and its spatial gradient for ray tracing. The refractive field $\eta$ is given as: 
\begin{equation} 
    \eta(\mathbf{x}) = \textbf{interp}(\sigma(G - 5) * \tau + 1, \mathbf{x})
\end{equation}
where $\sigma$ is the soft-plus function output, G is the sampled Gaussian Process output, and $\tau$ is the scaling factor set to control the magnitude of the RI field. For Section 4.1 in the paper, we set $\tau$ to be 1/50, 1/50, and 1/90 for the first, second and third rows respectively. 

\section{Refractive Radiative Transfer Equation}

\begin{figure*}
    \centering
    \includegraphics[width=0.7\textwidth]{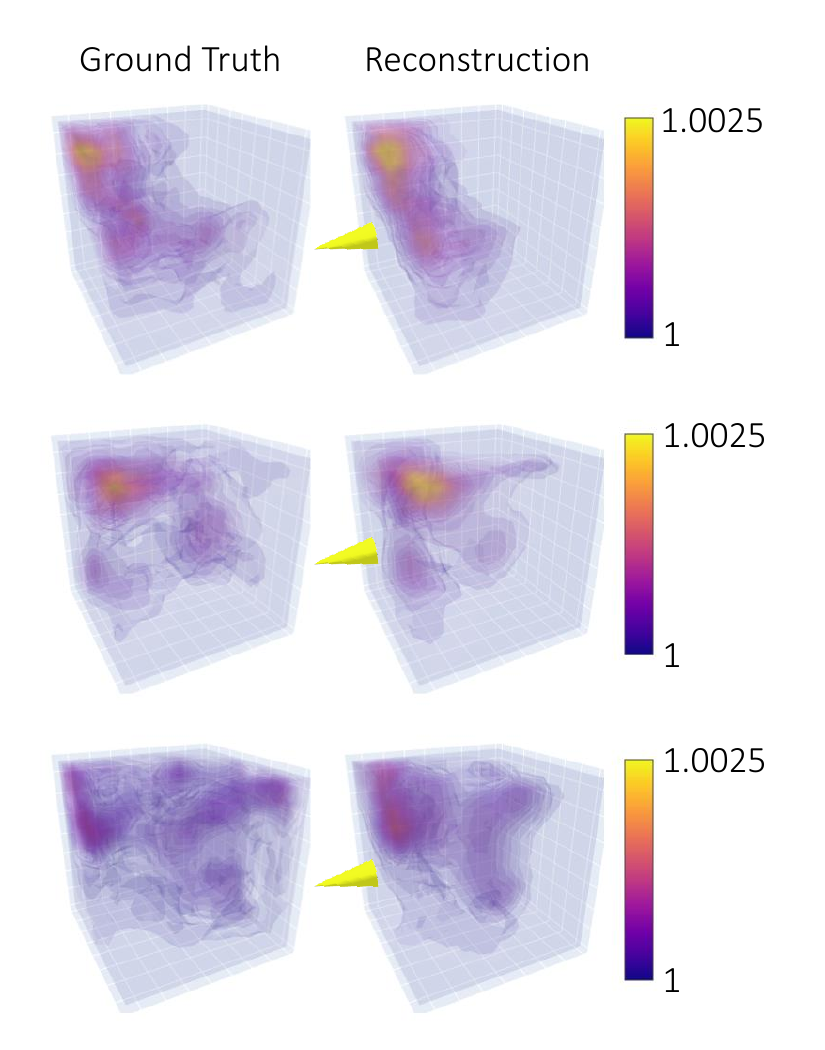}
    \caption{\textbf{Recovery with the RRTE}: In many applications the RRTE forward model \cite{ament2014refractive} more accurately describes how pixel intensities in the image plane should be calculated. Our method also works well in these cases. Yellow cones indicate sensor direction.}
    \label{fig:rrte}
\end{figure*}

In our work we calculate the intensity of a given image pixel by first solving for refracted ray paths with backwards ray-tracing then integrating emission coefficients along the traced ray path according to the Radiative Transfer Equation (RTE) \cite{chandrasekhar2013radiative}. This equation (Eq. 4 in the paper) relies on the fact that specific intensity $I$, otherwise known as radiance, is conserved along a ray bundle traveling in free space. However, for many applications of refractive tomography, this is not exactly true; instead, a quantity called basic radiance, equal to $\frac{I}{\eta^2}$, is conserved along a ray path, according to the Refractive Radiative Transfer Equation (RRTE) \cite{ament2014refractive}. Our method also works for this forward model; recovery results for the experiment given in paper Figure 4 but with the RRTE forward model are shown in Fig. \ref{fig:rrte}.

In the regime of gravitational lensing, we approximate the effects of general relativity on rays of light as refraction. We make the simplifying assumptions that we are imaging a stationary refractive field, and that the effects of redshift on emitted intensities are negligible; similar assumptions have been made in previous works \cite{levis2022gravitationally}. Under these assumptions, the regular RTE is sufficient to accurately calculate image pixel intensities \cite{vincent2011gyoto, rybicki1991radiative} ; thus, experiments in the main paper are performed with the RTE forward model.

{
    \small
    \bibliographystyle{ieeenat_fullname}
    \bibliography{supplement}
}

%% file: preamble.tex
\usepackage[dvipsnames]{xcolor}


%% file: main.bbl
\begin{thebibliography}{40}
\providecommand{\natexlab}[1]{#1}
\providecommand{\url}[1]{\texttt{#1}}
\expandafter\ifx\csname urlstyle\endcsname\relax
  \providecommand{\doi}[1]{doi: #1}\else
  \providecommand{\doi}{doi: \begingroup \urlstyle{rm}\Url}\fi

\bibitem[Alterman et~al.(2014{\natexlab{a}})Alterman, Schechner, Vo, and Narasimhan]{alterman2014passive}
Marina Alterman, Yoav~Y Schechner, Minh Vo, and Srinivasa~G Narasimhan.
\newblock Passive tomography of turbulence strength.
\newblock In \emph{Computer Vision--ECCV 2014: 13th European Conference, Zurich, Switzerland, September 6-12, 2014, Proceedings, Part IV 13}, pages 47--60. Springer, 2014{\natexlab{a}}.

\bibitem[Alterman et~al.(2014{\natexlab{b}})Alterman, Swirski, and Schechner]{alterman2014stella}
Marina Alterman, Yohay Swirski, and Yoav~Y Schechner.
\newblock Stella maris: Stellar marine refractive imaging sensor.
\newblock In \emph{2014 IEEE International Conference on Computational Photography (ICCP)}, pages 1--10. IEEE, 2014{\natexlab{b}}.

\bibitem[Ament et~al.(2014)Ament, Bergmann, and Weiskopf]{ament2014refractive}
Marco Ament, Christoph Bergmann, and Daniel Weiskopf.
\newblock Refractive radiative transfer equation.
\newblock \emph{ACM Transactions on Graphics (TOG)}, 33\penalty0 (2):\penalty0 1--22, 2014.

\bibitem[Atcheson et~al.(2008)Atcheson, Ihrke, Heidrich, Tevs, Bradley, Magnor, and Seidel]{atcheson2008time}
Bradley Atcheson, Ivo Ihrke, Wolfgang Heidrich, Art Tevs, Derek Bradley, Marcus Magnor, and Hans-Peter Seidel.
\newblock Time-resolved 3d capture of non-stationary gas flows.
\newblock \emph{ACM transactions on graphics (TOG)}, 27\penalty0 (5):\penalty0 1--9, 2008.

\bibitem[Bartelmann and Schneider(2001)]{bartelmann2001weak}
Matthias Bartelmann and Peter Schneider.
\newblock Weak gravitational lensing.
\newblock \emph{Physics Reports}, 340\penalty0 (4-5):\penalty0 291--472, 2001.

\bibitem[Bemana et~al.(2022)Bemana, Myszkowski, Revall~Frisvad, Seidel, and Ritschel]{bemana2022eikonal}
Mojtaba Bemana, Karol Myszkowski, Jeppe Revall~Frisvad, Hans-Peter Seidel, and Tobias Ritschel.
\newblock Eikonal fields for refractive novel-view synthesis.
\newblock In \emph{ACM SIGGRAPH 2022 Conference Proceedings}, pages 1--9, 2022.

\bibitem[Bradbury et~al.(2018)Bradbury, Frostig, Hawkins, Johnson, Leary, Maclaurin, Necula, Paszke, Vander{P}las, Wanderman-{M}ilne, and Zhang]{jax2018github}
James Bradbury, Roy Frostig, Peter Hawkins, Matthew~James Johnson, Chris Leary, Dougal Maclaurin, George Necula, Adam Paszke, Jake Vander{P}las, Skye Wanderman-{M}ilne, and Qiao Zhang.
\newblock {JAX}: composable transformations of {P}ython+{N}um{P}y programs, 2018.

\bibitem[Bridson(2007)]{bridson2007fast}
Robert Bridson.
\newblock Fast poisson disk sampling in arbitrary dimensions.
\newblock \emph{SIGGRAPH sketches}, 10\penalty0 (1):\penalty0 1, 2007.

\bibitem[Chen et~al.(2018)Chen, Rubanova, Bettencourt, and Duvenaud]{chen2018neural}
Ricky~TQ Chen, Yulia Rubanova, Jesse Bettencourt, and David~K Duvenaud.
\newblock Neural ordinary differential equations.
\newblock \emph{Advances in neural information processing systems}, 31, 2018.

\bibitem[Chen and Zhang(2019)]{chen2019learning}
Zhiqin Chen and Hao Zhang.
\newblock Learning implicit fields for generative shape modeling.
\newblock In \emph{Proceedings of the IEEE/CVF Conference on Computer Vision and Pattern Recognition}, pages 5939--5948, 2019.

\bibitem[Choi et~al.(2007)Choi, Fang-Yen, Badizadegan, Oh, Lue, Dasari, and Feld]{choi2007tomographic}
Wonshik Choi, Christopher Fang-Yen, Kamran Badizadegan, Seungeun Oh, Niyom Lue, Ramachandra~R Dasari, and Michael~S Feld.
\newblock Tomographic phase microscopy.
\newblock \emph{Nature methods}, 4\penalty0 (9):\penalty0 717--719, 2007.

\bibitem[Chowdhury et~al.(2019)Chowdhury, Chen, Eckert, Ren, Wu, Repina, and Waller]{chowdhury2019high}
Shwetadwip Chowdhury, Michael Chen, Regina Eckert, David Ren, Fan Wu, Nicole Repina, and Laura Waller.
\newblock High-resolution 3d refractive index microscopy of multiple-scattering samples from intensity images.
\newblock \emph{Optica}, 6\penalty0 (9):\penalty0 1211--1219, 2019.

\bibitem[Dormand and Prince(1980)]{dormand1980family}
John~R Dormand and Peter~J Prince.
\newblock A family of embedded runge-kutta formulae.
\newblock \emph{Journal of computational and applied mathematics}, 6\penalty0 (1):\penalty0 19--26, 1980.

\bibitem[He et~al.(2015)He, Zhang, Ren, and Sun]{he2015delving}
Kaiming He, Xiangyu Zhang, Shaoqing Ren, and Jian Sun.
\newblock Delving deep into rectifiers: Surpassing human-level performance on imagenet classification.
\newblock In \emph{Proceedings of the IEEE international conference on computer vision}, pages 1026--1034, 2015.

\bibitem[Hoekstra and Jain(2008)]{hoekstra2008weak}
Henk Hoekstra and Bhuvnesh Jain.
\newblock Weak gravitational lensing and its cosmological applications.
\newblock \emph{Annual Review of Nuclear and Particle Science}, 58:\penalty0 99--123, 2008.

\bibitem[Ihrke et~al.(2007)Ihrke, Ziegler, Tevs, Theobalt, Magnor, and Seidel]{ihrke2007eikonal}
Ivo Ihrke, Gernot Ziegler, Art Tevs, Christian Theobalt, Marcus Magnor, and Hans-Peter Seidel.
\newblock Eikonal rendering: Efficient light transport in refractive objects.
\newblock \emph{ACM Transactions on Graphics (TOG)}, 26\penalty0 (3):\penalty0 59--es, 2007.

\bibitem[Kingma and Ba(2014)]{kingma2014adam}
Diederik~P Kingma and Jimmy Ba.
\newblock Adam: A method for stochastic optimization.
\newblock \emph{arXiv preprint arXiv:1412.6980}, 2014.

\bibitem[Laureijs et~al.(2011)Laureijs, Amiaux, Arduini, Augueres, Brinchmann, Cole, Cropper, Dabin, Duvet, Ealet, et~al.]{laureijs2011euclid}
Rene Laureijs, J Amiaux, S Arduini, J-L Augueres, J Brinchmann, R Cole, M Cropper, C Dabin, L Duvet, A Ealet, et~al.
\newblock Euclid definition study report.
\newblock \emph{arXiv preprint arXiv:1110.3193}, 2011.

\bibitem[Leonard et~al.(2012)Leonard, Dup{\'e}, and Starck]{leonard2012compressed}
Adrienne Leonard, F-X Dup{\'e}, and J-L Starck.
\newblock A compressed sensing approach to 3d weak lensing.
\newblock \emph{Astronomy \& Astrophysics}, 539:\penalty0 A85, 2012.

\bibitem[Leonard et~al.(2014)Leonard, Lanusse, and Starck]{leonard2014glimpse}
Adrienne Leonard, Fran{\c{c}}ois Lanusse, and Jean-Luc Starck.
\newblock Glimpse: accurate 3d weak lensing reconstructions using sparsity.
\newblock \emph{Monthly Notices of the Royal Astronomical Society}, 440\penalty0 (2):\penalty0 1281--1294, 2014.

\bibitem[Levis et~al.(2022)Levis, Srinivasan, Chael, Ng, and Bouman]{levis2022gravitationally}
Aviad Levis, Pratul~P Srinivasan, Andrew~A Chael, Ren Ng, and Katherine~L Bouman.
\newblock Gravitationally lensed black hole emission tomography.
\newblock In \emph{Proceedings of the IEEE/CVF Conference on Computer Vision and Pattern Recognition}, pages 19841--19850, 2022.

\bibitem[Li et~al.(2021)Li, Hou, Wang, Tan, Liu, and Zhang]{li2021end}
Zongling Li, Qingyu Hou, Zhipeng Wang, Fanjiao Tan, Jin Liu, and Wei Zhang.
\newblock End-to-end learned single lens design using fast differentiable ray tracing.
\newblock \emph{Optics Letters}, 46\penalty0 (21):\penalty0 5453--5456, 2021.

\bibitem[Lin et~al.(2009)Lin, Ritzwoller, and Snieder]{lin2009eikonal}
Fan-Chi Lin, Michael~H Ritzwoller, and Roel Snieder.
\newblock Eikonal tomography: surface wave tomography by phase front tracking across a regional broad-band seismic array.
\newblock \emph{Geophysical Journal International}, 177\penalty0 (3):\penalty0 1091--1110, 2009.

\bibitem[Liu et~al.(2022)Liu, Sun, Zhu, Tian, and Kamilov]{liu2022recovery}
Renhao Liu, Yu Sun, Jiabei Zhu, Lei Tian, and Ulugbek~S Kamilov.
\newblock Recovery of continuous 3d refractive index maps from discrete intensity-only measurements using neural fields.
\newblock \emph{Nature Machine Intelligence}, 4\penalty0 (9):\penalty0 781--791, 2022.

\bibitem[Mildenhall et~al.(2021)Mildenhall, Srinivasan, Tancik, Barron, Ramamoorthi, and Ng]{mildenhall2021nerf}
Ben Mildenhall, Pratul~P Srinivasan, Matthew Tancik, Jonathan~T Barron, Ravi Ramamoorthi, and Ren Ng.
\newblock Nerf: Representing scenes as neural radiance fields for view synthesis.
\newblock \emph{Communications of the ACM}, 65\penalty0 (1):\penalty0 99--106, 2021.

\bibitem[Narayan and Bartelmann(1996)]{narayan1996lectures}
Ramesh Narayan and Matthias Bartelmann.
\newblock Lectures on gravitational lensing.
\newblock \emph{arXiv preprint astro-ph/9606001}, 1996.

\bibitem[Nelson et~al.(2015)Nelson, Pillepich, Genel, Vogelsberger, Springel, Torrey, Rodriguez-Gomez, Sijacki, Snyder, Griffen, et~al.]{nelson2015illustris}
Dylan Nelson, Annalisa Pillepich, Shy Genel, Mark Vogelsberger, Volker Springel, Paul Torrey, Vicente Rodriguez-Gomez, Debora Sijacki, Gregory~F Snyder, Brendan Griffen, et~al.
\newblock The illustris simulation: Public data release.
\newblock \emph{Astronomy and Computing}, 13:\penalty0 12--37, 2015.

\bibitem[Pediredla et~al.(2020)Pediredla, Chalmiani, Scopelliti, Chamanzar, Narasimhan, and Gkioulekas]{pediredla2020path}
Adithya Pediredla, Yasin~Karimi Chalmiani, Matteo~Giuseppe Scopelliti, Maysamreza Chamanzar, Srinivasa Narasimhan, and Ioannis Gkioulekas.
\newblock Path tracing estimators for refractive radiative transfer.
\newblock \emph{ACM Transactions on Graphics (TOG)}, 39\penalty0 (6):\penalty0 1--15, 2020.

\bibitem[Pillepich et~al.(2018)Pillepich, Springel, Nelson, Genel, Naiman, Pakmor, Hernquist, Torrey, Vogelsberger, Weinberger, et~al.]{pillepich2018simulating}
Annalisa Pillepich, Volker Springel, Dylan Nelson, Shy Genel, Jill Naiman, R{\"u}diger Pakmor, Lars Hernquist, Paul Torrey, Mark Vogelsberger, Rainer Weinberger, et~al.
\newblock Simulating galaxy formation with the illustristng model.
\newblock \emph{Monthly Notices of the Royal Astronomical Society}, 473\penalty0 (3):\penalty0 4077--4106, 2018.

\bibitem[Pontryagin(1987)]{pontryagin1987mathematical}
Lev~Semenovich Pontryagin.
\newblock \emph{Mathematical theory of optimal processes}.
\newblock CRC press, 1987.

\bibitem[Reed et~al.(2002)Reed, Yan, and Schnitzer]{reed2002gradient}
William~A Reed, Man~F Yan, and Mark~J Schnitzer.
\newblock Gradient-index fiber-optic microprobes for minimally invasive in vivo low-coherence interferometry.
\newblock \emph{Optics letters}, 27\penalty0 (20):\penalty0 1794--1796, 2002.

\bibitem[Schneider et~al.()Schneider, Ehlers, Falco, et~al.]{schneidergravitational}
Peter Schneider, J{\"u}rgen Ehlers, Emilio~E Falco, et~al.
\newblock Gravitational lenses [electronic resource].

\bibitem[Simon et~al.(2009)Simon, Taylor, and Hartlap]{simon2009unfolding}
Patrick Simon, AN Taylor, and Jan Hartlap.
\newblock Unfolding the matter distribution using three-dimensional weak gravitational lensing.
\newblock \emph{Monthly Notices of the Royal Astronomical Society}, 399\penalty0 (1):\penalty0 48--68, 2009.

\bibitem[Sung et~al.(2009)Sung, Choi, Fang-Yen, Badizadegan, Dasari, and Feld]{sung2009optical}
Yongjin Sung, Wonshik Choi, Christopher Fang-Yen, Kamran Badizadegan, Ramachandra~R Dasari, and Michael~S Feld.
\newblock Optical diffraction tomography for high resolution live cell imaging.
\newblock \emph{Optics express}, 17\penalty0 (1):\penalty0 266--277, 2009.

\bibitem[Takada(2010)]{takada2010subaru}
Masahiro Takada.
\newblock Subaru hyper suprime-cam project.
\newblock In \emph{AIP Conference Proceedings}, pages 120--127. American Institute of Physics, 2010.

\bibitem[Tancik et~al.(2020)Tancik, Srinivasan, Mildenhall, Fridovich-Keil, Raghavan, Singhal, Ramamoorthi, Barron, and Ng]{tancik2020fourier}
Matthew Tancik, Pratul Srinivasan, Ben Mildenhall, Sara Fridovich-Keil, Nithin Raghavan, Utkarsh Singhal, Ravi Ramamoorthi, Jonathan Barron, and Ren Ng.
\newblock Fourier features let networks learn high frequency functions in low dimensional domains.
\newblock \emph{Advances in Neural Information Processing Systems}, 33:\penalty0 7537--7547, 2020.

\bibitem[Teh et~al.(2022)Teh, O'Toole, and Gkioulekas]{teh2022adjoint}
Arjun Teh, Matthew O'Toole, and Ioannis Gkioulekas.
\newblock Adjoint nonlinear ray tracing.
\newblock \emph{ACM Transactions on Graphics (TOG)}, 41\penalty0 (4):\penalty0 1--13, 2022.

\bibitem[Tseng et~al.(2021)Tseng, Mosleh, Mannan, St-Arnaud, Sharma, Peng, Braun, Nowrouzezahrai, Lalonde, and Heide]{tseng2021differentiable}
Ethan Tseng, Ali Mosleh, Fahim Mannan, Karl St-Arnaud, Avinash Sharma, Yifan Peng, Alexander Braun, Derek Nowrouzezahrai, Jean-Francois Lalonde, and Felix Heide.
\newblock Differentiable compound optics and processing pipeline optimization for end-to-end camera design.
\newblock \emph{ACM Transactions on Graphics (TOG)}, 40\penalty0 (2):\penalty0 1--19, 2021.

\bibitem[VanderPlas et~al.(2011)VanderPlas, Connolly, Jain, and Jarvis]{vanderplas2011three}
JT VanderPlas, AJ Connolly, Bhuvnesh Jain, and Mike Jarvis.
\newblock Three-dimensional reconstruction of the density field: An svd approach to weak-lensing tomography.
\newblock \emph{The Astrophysical Journal}, 727\penalty0 (2):\penalty0 118, 2011.

\bibitem[Xue et~al.(2014)Xue, Rubinstein, Wadhwa, Levin, Durand, and Freeman]{xue2014refraction}
Tianfan Xue, Michael Rubinstein, Neal Wadhwa, Anat Levin, Fredo Durand, and William~T Freeman.
\newblock Refraction wiggles for measuring fluid depth and velocity from video.
\newblock In \emph{Computer Vision--ECCV 2014: 13th European Conference, Zurich, Switzerland, September 6-12, 2014, Proceedings, Part III 13}, pages 767--782. Springer, 2014.

\end{thebibliography}


\begin{thebibliography}{9}
\providecommand{\natexlab}[1]{#1}
\providecommand{\url}[1]{\texttt{#1}}
\expandafter\ifx\csname urlstyle\endcsname\relax
  \providecommand{\doi}[1]{doi: #1}\else
  \providecommand{\doi}{doi: \begingroup \urlstyle{rm}\Url}\fi

\bibitem[Ament et~al.(2014)Ament, Bergmann, and Weiskopf]{ament2014refractive}
Marco Ament, Christoph Bergmann, and Daniel Weiskopf.
\newblock Refractive radiative transfer equation.
\newblock \emph{ACM Transactions on Graphics (TOG)}, 33\penalty0 (2):\penalty0 1--22, 2014.

\bibitem[Atcheson et~al.(2008)Atcheson, Ihrke, Heidrich, Tevs, Bradley, Magnor, and Seidel]{atcheson2008time}
Bradley Atcheson, Ivo Ihrke, Wolfgang Heidrich, Art Tevs, Derek Bradley, Marcus Magnor, and Hans-Peter Seidel.
\newblock Time-resolved 3d capture of non-stationary gas flows.
\newblock \emph{ACM transactions on graphics (TOG)}, 27\penalty0 (5):\penalty0 1--9, 2008.

\bibitem[Chandrasekhar(2013)]{chandrasekhar2013radiative}
Subrahmanyan Chandrasekhar.
\newblock \emph{Radiative transfer}.
\newblock Courier Corporation, 2013.

\bibitem[Levis et~al.(2022)Levis, Srinivasan, Chael, Ng, and Bouman]{levis2022gravitationally}
Aviad Levis, Pratul~P Srinivasan, Andrew~A Chael, Ren Ng, and Katherine~L Bouman.
\newblock Gravitationally lensed black hole emission tomography.
\newblock In \emph{Proceedings of the IEEE/CVF Conference on Computer Vision and Pattern Recognition}, pages 19841--19850, 2022.

\bibitem[Mandelbaum et~al.(2018)Mandelbaum, Miyatake, Hamana, Oguri, Simet, Armstrong, Bosch, Murata, Lanusse, Leauthaud, et~al.]{mandelbaum2018first}
Rachel Mandelbaum, Hironao Miyatake, Takashi Hamana, Masamune Oguri, Melanie Simet, Robert Armstrong, James Bosch, Ryoma Murata, Fran{\c{c}}ois Lanusse, Alexie Leauthaud, et~al.
\newblock The first-year shear catalog of the subaru hyper suprime-cam subaru strategic program survey.
\newblock \emph{Publications of the Astronomical Society of Japan}, 70\penalty0 (SP1):\penalty0 S25, 2018.

\bibitem[Oguri et~al.(2018)Oguri, Miyazaki, Hikage, Mandelbaum, Utsumi, Miyatake, Takada, Armstrong, Bosch, Komiyama, et~al.]{oguri2018two}
Masamune Oguri, Satoshi Miyazaki, Chiaki Hikage, Rachel Mandelbaum, Yousuke Utsumi, Hironao Miyatake, Masahiro Takada, Robert Armstrong, James Bosch, Yutaka Komiyama, et~al.
\newblock Two-and three-dimensional wide-field weak lensing mass maps from the hyper suprime-cam subaru strategic program s16a data.
\newblock \emph{Publications of the Astronomical Society of Japan}, 70\penalty0 (SP1):\penalty0 S26, 2018.

\bibitem[Rybicki and Lightman(1991)]{rybicki1991radiative}
George~B Rybicki and Alan~P Lightman.
\newblock \emph{Radiative processes in astrophysics}.
\newblock John Wiley \& Sons, 1991.

\bibitem[Teh et~al.(2022)Teh, O'Toole, and Gkioulekas]{teh2022adjoint}
Arjun Teh, Matthew O'Toole, and Ioannis Gkioulekas.
\newblock Adjoint nonlinear ray tracing.
\newblock \emph{ACM Transactions on Graphics (TOG)}, 41\penalty0 (4):\penalty0 1--13, 2022.

\bibitem[Vincent et~al.(2011)Vincent, Paumard, Gourgoulhon, and Perrin]{vincent2011gyoto}
Frederic~H Vincent, Thibaut Paumard, Eric Gourgoulhon, and Guy Perrin.
\newblock Gyoto: a new general relativistic ray-tracing code.
\newblock \emph{Classical and Quantum Gravity}, 28\penalty0 (22):\penalty0 225011, 2011.

\end{thebibliography}
